\documentclass{IOS-Book-Article}

\usepackage{mathptmx}
\usepackage{soul}\setuldepth{article}

\usepackage{graphicx}
\usepackage{algorithm} 
\usepackage{algpseudocode} 
\usepackage{listings}
\usepackage[
singlelinecheck=false 
]{caption}
\usepackage{subcaption}

\def\hb{\hbox to 11.5 cm{}}

\begin{document}

\pagestyle{headings}
\def\thepage{}
\begin{frontmatter}              

\title{The Effectiveness of Bidirectional Generative Patent Language Models}

\markboth{}{September 2022\hb}

\author{\fnms{Jieh-Sheng} \snm{Lee}\orcid{0000-0002-0990-6170}%
\thanks{Assistant Professor of Law. Ph.D. in Computer Science. Admitted in New York. Email: jasonlee@nycu.edu.tw}},

\address[A]{National Yang Ming Chiao Tung University School of Law}

\begin{abstract}
Generative patent language models can assist humans to write patent text more effectively. The question is how to measure effectiveness from a human-centric perspective and how to improve effectiveness. In this manuscript, a simplified design of the autocomplete function is proposed to increase effectiveness by more than 10\%. With the new design, the effectiveness of autocomplete can reach more than 60\%, which means that more than 60\% of keystrokes can be saved by autocomplete. Since writing patent text does not necessarily start from the beginning to the end, a question is whether the generative model can assist a user no matter where to start writing. To answer the question, the generative models in this manuscript are pre-trained with training data in both directions. The generative models become bidirectional. Since text generation is bidirectional, the calculation of autocomplete effectiveness can be bidirectional and starts from anywhere in the text. After thorough experiments, a key finding is that the autocomplete effectiveness of a model for the same text remains similar no matter where the calculation starts. The finding indicates that such bidirectional models can assist a user at a similar level, no matter where the user starts to write. 
\end{abstract}

\end{frontmatter}
\markboth{September 2022\hb}{September 2022\hb}

\section{Introduction}
\label{section:introduction}

Large language models have achieved notable success in natural language generation (NLG) tasks in recent years. Until now, very few language models have been dedicated to the patent domain. Furthermore, most language models are autoregressive by predicting the \emph{next} token after having read all the previous ones. Very few language models work backward by predicting the \emph{previous} token. In this manuscript, the author pre-trained and fine-tuned large language models dedicated to the patent domain. The model can predict the previous token, in addition to predicting the next token.
Predicting the previous token is implemented by reversing the tokens for both inputs and outputs. The training dataset also contains tokenized sequences in both forward and backward directions.
By doing so, the patent-specific language models in this manuscript can generate patent text in both forward and backward directions.  The motivation is to make patent text generation flexible because human writing does not necessarily start from the beginning to the end. Assuming that the thought process in human's mind is a back-and-forth process, a bidirectional generative language model should assist humans more flexibly. 

To evaluate the performance of generative patent language models, this manuscript follows the Autocomplete Effectiveness (AE) ratio proposed in~\cite{jiehsheng_n01}. The ratio is used to measure how many keystrokes can be saved for a user if an autocomplete function is provided and based on the generative model. The higher the AE ratio, the more the keystrokes are saved. The details in~\cite{jiehsheng_n01} is provided in the next section. The contributions of this manuscript include: (1) proposed a simplified version of the autocomplete function in pseudocode to reach higher AE ratios, (2) making the calculation of AE ratios bidirectional, (3) observed similar AE ratios when using different starting positions of the text for calculation, (4) fine-tuned the language models with a specific CPC Subclass for measuring the improvement of AE ratios, and (5) released the models, datasets, experiment results, and sample code in this manuscript. 

\section{Related Work}
\label{section:related_work}
This manuscript is the follow-up work of~\cite{jiehsheng_n01}, which proposed the AE ratio to evaluate generative language models. 
In~\cite{jiehsheng_n01}, the patent language model is called PatentGPT-J. The PatentGPT-J models are based on the GPT-J-6B~\cite{gpt-j-github} models and pre-trained with a patent corpus from scratch. 
The use of a Transformer~\cite{Transformer} language model for patent text generation was first proposed in~\cite{jiehsheng03} by fine-tuning a GPT-2~\cite{gpt2_Radrof01} model with patent corpus. 
A Transformer model is a deep learning model that adopts the mechanism of self-attention and learns context by tracking relationships in sequential data. 
The idea of generating patent text backward was introduced in~\cite{jiehsheng05}. The research in~\cite{jiehsheng05} focuses on controlling patent text generation by structural metadata. The effective way to generate text backward and how to measure effectiveness were not addressed in~\cite{jiehsheng05}. In~\cite{jiehsheng04}, the idea of making a patent language model personalized was introduced but not implemented. The hypothesis in~\cite{jiehsheng04} is that, by fine-tuning the model with a specific kind of patent data, the model can be personalized and perform better. In this manuscript, the pre-trained PatentGPT-J models are fine-tuned with a specific CPC Subclass to address the hypothesis. Another work related to patent text generation is relevant to prior art search and is described in~\cite{jiehsheng09}. Except for these works, patent text generation remains a niche research topic less explored.

It should be noted that the training data for GPT-J-6B~\cite{gpt-j-github} contains the USPTO Backgrounds dataset. The data set contains the background sections extracted from patents granted by the United States Patent and Trademark Office (USPTO). The coverage of the dataset ranges from 1976 to 2020. The initial tests by the author of this manuscript show that the GPT-J-6B model is capable of generating patent text. The script to collect the USPTO Backgrounds dataset is available at~\cite{EleutherAI_pile_uspto}. The same script is used in the BigScience research workshop, according to~\cite{BigScienceCorpus_pile_uspto}. The workshop released the BigScience Large Open-science Open-access Multilingual Language Model (BLOOM) model. The BLOOM model is an autoregressive language model based on the GPT-3~\cite{GPT-3-NEURIPS2020} architecture. With its 176 billion parameters, BLOOM is capable of generating text in 46 natural languages and 13 programming languages. At the time of this writing, the model is the largest publicly available open and multilingual model. Initial tests show that the BLOOM model is also capable of generating patent text. In terms of performance, it is hypothesized that the PatentGPT-J models in this manuscript will perform better than both the GPT-J-6B model and the BLOOM model.  PatentGPT-J models should have higher AE ratios because the training data contain patents only and cover more sections of patent documents. The validation of this hypothesis is a research topic of its own in the future. 



\section{Implementation}
\label{section:implementation}

Based on~\cite{jiehsheng_n01}, the major implementations in this manuscript include: (1) making the autocomplete function simpler and the AE ratio higher, (2) generating patent text backward, and (3) fine-tuning the models with CPC Subclass G06N. These implementations are explained as follows.

\subsection{Simpler Autocomplete Function}
\label{subsection:simpler_autocomplete}

The purpose of the AE ratio is to evaluate a language model from a human-centric perspective: how many manual keystrokes can be saved by the autocomplete function based on the generative language model?
A higher AE ratio means that the autocomplete function works more effectively and more manual keystrokes are saved. 
This section describes how the autocomplete function can be simplified to obtain a higher AE ratio. 
The original AE ratio and the implementations of the PatentGPT-J models are described in~\cite{jiehsheng_n01}. 
The improvement in AE ratio is achieved by simplifying the user interface (UI). In the conceptual UI design in~\cite{jiehsheng_n01}, a user has to press the ``downarrow'' ($\downarrow$) key and the ``tab'' key to complete the autocomplete function. The user selects a preferred token in the top 10 tokens predicted by the model in this way. Such UI and user's operation is common as an input method. In this manuscript, a simpler UI is proposed using keys 0$\sim$9 to represent the top 10 tokens. Therefore, the user can select a token by pressing only one key instead of multiple keystrokes. 
For example, in the previous design in~\cite{jiehsheng_n01}, in order to select the 6th token, the user has to press the ``downarrow'' ($\downarrow$) key five times and the ``tab'' key once. Six keystrokes are required. In the new and simplified design in this manuscript, the user can press the ``5'' key to select the 6th token of the top 10 tokens. Five keystrokes are saved (``0'' key to represent the first token). The pseudocode in Algorithm~\ref{alg:counting_keystrokes_v2} shows the simplified design to calculate the minimal number of keystrokes with autocomplete. Lines 11-12 in Algorithm~\ref{alg:counting_keystrokes_v2} are the key difference compared to the previous pseudocode in~\cite{jiehsheng_n01}. Lines 11-12 provide using one key to select a top-10 token. 

\begin{algorithm}
  \caption{Calculating AE Ratio}
  \label{alg:counting_keystrokes_v2} 
  \begin{algorithmic}[1]
    \State {$patent\_text$} $\leftarrow$ read one new patent claim
    \State {$patent\_tokens$} $\leftarrow$ tokenizer.encode({$patent\_text$})  
    \State $keys\_auto$ $\leftarrow$ 0
    \State $keys\_manual$ $\leftarrow$ 0
    \For {$i=1$ to len($patent\_tokens$)-1}
      \State {$prompt$} $\leftarrow$ $patent\_tokens[:i]$
      \State {$next\_token$} $\leftarrow$ $patent\_tokens[i]$
      \State {$next\_text$} $\leftarrow$ tokenizer.decode($next\_token$)
      \State {$next\_text$} $\leftarrow$ {$next\_text$}.strip() \Comment{remove space}
      \State $top$ $10$ $tokens$ $\leftarrow$ model.predict($prompt$)
      \If {$next\_token$ in [$top$ $10$ $tokens$] }
        \State $keys\_auto$ += 1 \Comment{autocomplete by ``0 $\sim$ 9'' key}
      \Else 
        \State $keys\_auto$ += len($next\_text$) \Comment{manual typing}
      \EndIf
      \State $keys\_manual$ += len($next\_text$) \Comment{manual typing}
    \EndFor
    \State $AE\_ratio$ = ($keys\_manual$ - $keys\_auto$) / $keys\_manual$
  \end{algorithmic} 
\end{algorithm}



\subsection{Back and Forth}
\label{subsection:back_and_forth}

\begin{figure}[h]
\centering
\begin{subfigure}{.5\textwidth}
  \centering
  \includegraphics[width=0.9\textwidth, keepaspectratio]{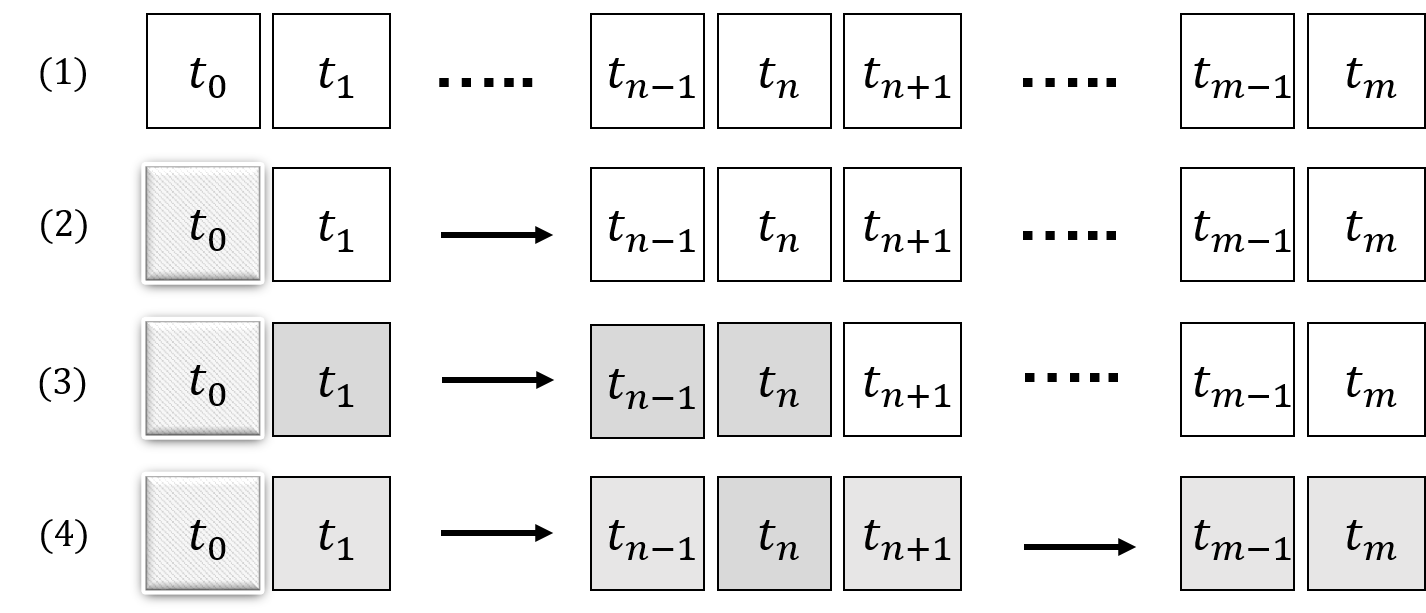}
  \captionsetup{justification=centering}
  \caption{Forward \& From the Beginning}
  \label{fig_forward_only}
\end{subfigure}%
\begin{subfigure}{.5\textwidth}
  \centering
  \includegraphics[width=0.9\textwidth, keepaspectratio]{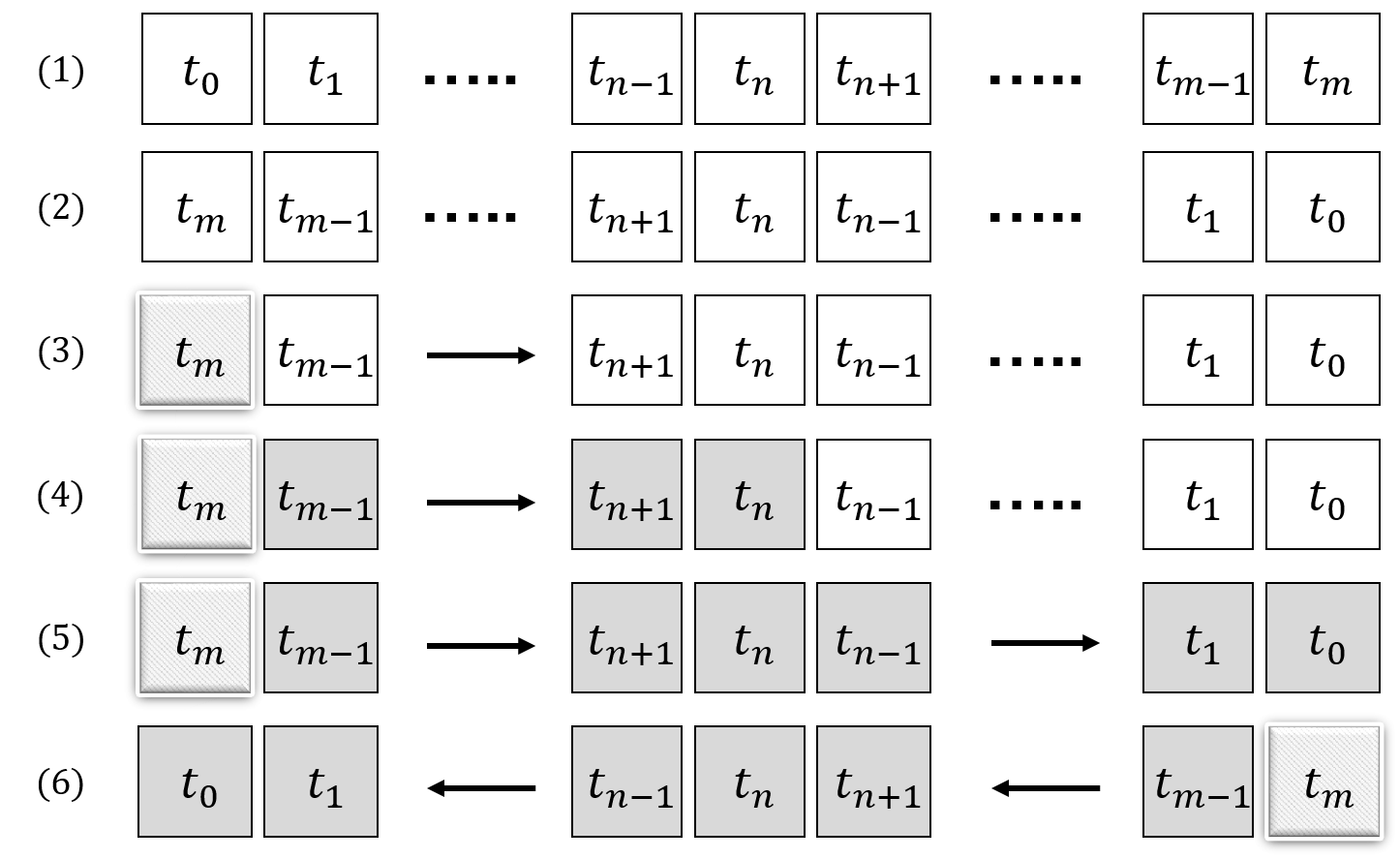}
  \captionsetup{justification=centering}
  \caption{Backward \& From the End}
  \label{fig_backward_only}
\end{subfigure}
\caption{Text Generation}
\label{fig:text_generation_1}
\end{figure}

The calculation of the previous AE ratio in~\cite{jiehsheng_n01} is defined as calculating forward to reach the end. The calculation starts from the beginning of the input text only. Fig.~\ref{fig_forward_only} shows how it works. Given some input text, the objective is to calculate the ratio of saved keystrokes. In Fig.~\ref{fig_forward_only}, row (1) shows the tokens of the input text after tokenization. Row (2) indicates that the calculation starts from \emph{$t_0$}. Row (3) indicates that the calculation moves forward. Row (4) shows that all tokens are calculated after reaching the end. The four steps in this use case are intuitive. The previous AE ratio in~\cite{jiehsheng_n01} addresses only this use case. 

Fig.~\ref{fig_backward_only} shows the first enhancement in this manuscript: calculating backward and starting from the end of the input text. In Fig.~\ref{fig_backward_only}, row (1) is the same, showing the tokens of the input text. Row (2) shows the reversed sequence of row (1). Row (3) indicates that the calculation starts from \emph{$t_m$} (the end of the original input text). Row (4) indicates that the calculation moves forward for the reversed tokens (backward in effect for the original tokens). Row (5) shows that all tokens are calculated. Row (6) shows the reversed sequence of row (5) to represent the calculation backward from the end of the original input text.


\begin{figure}[h]
\centering
\begin{subfigure}{.5\textwidth}
  \centering
  \includegraphics[width=0.9\textwidth, keepaspectratio]{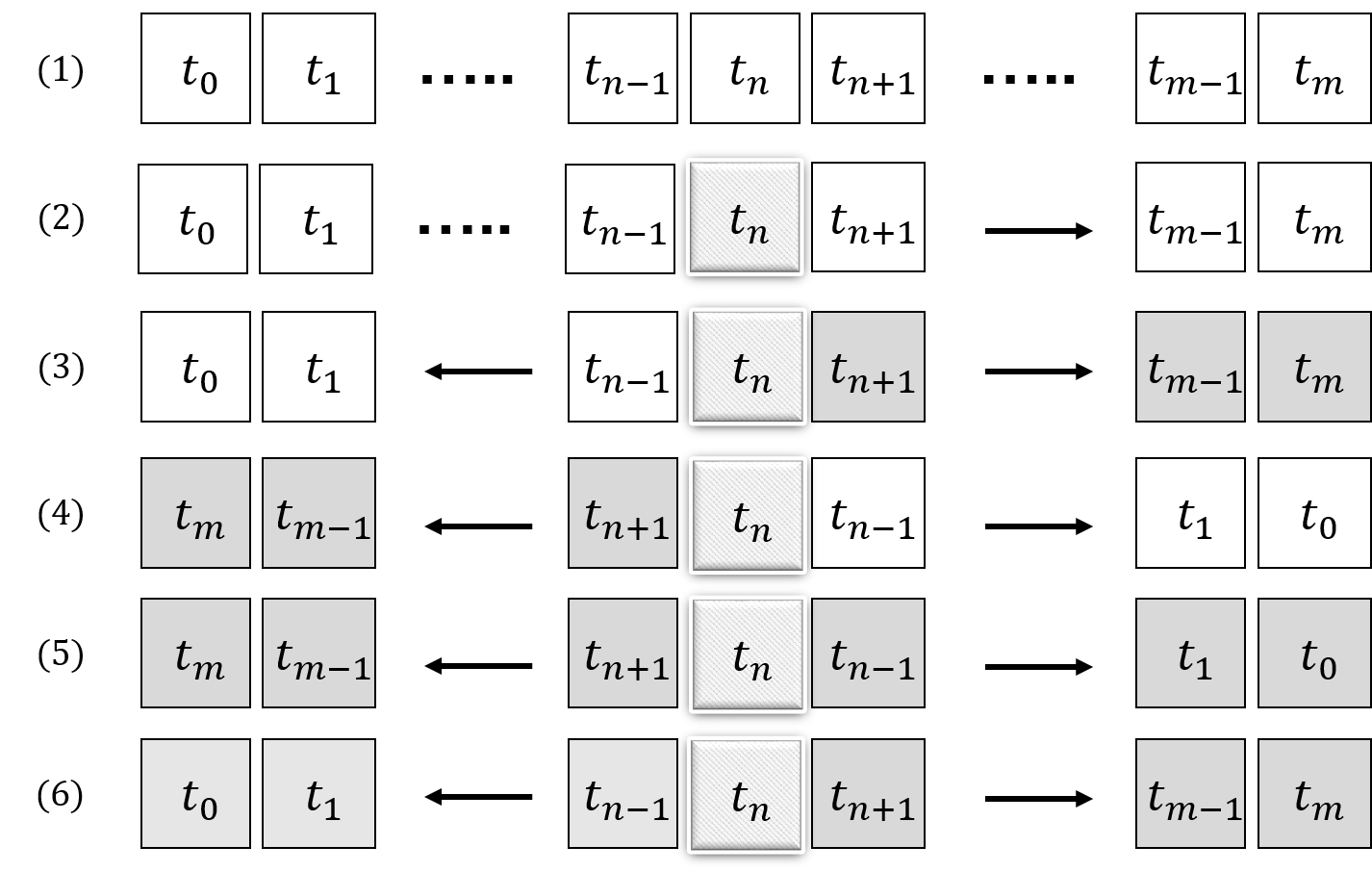}
  \captionsetup{justification=centering}
  \caption{Forward \& In the Middle}
  \label{fig_forward_in_the_middle}
\end{subfigure}%
\begin{subfigure}{.5\textwidth}
  \centering
  \includegraphics[width=0.9\textwidth, keepaspectratio]{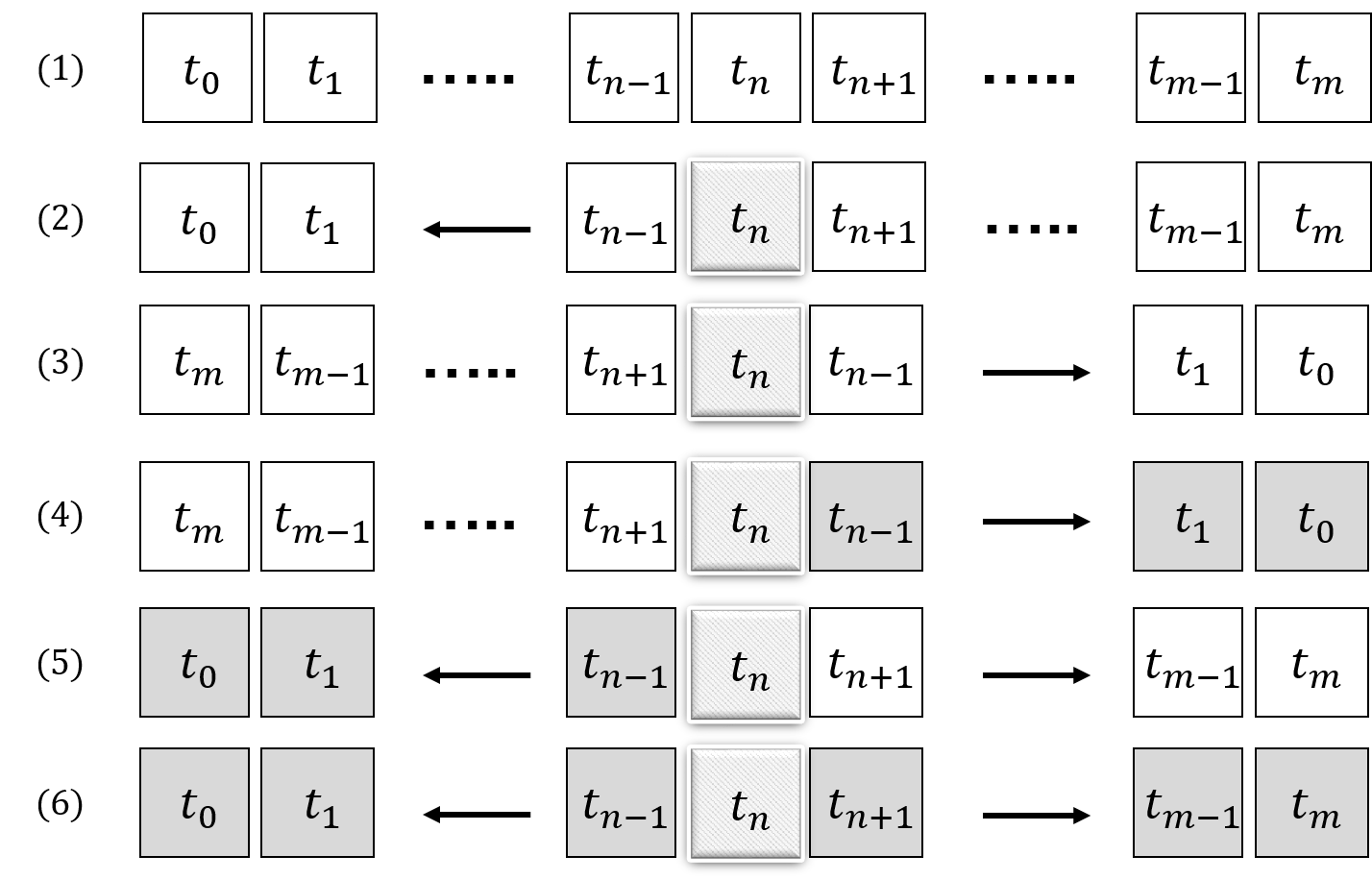}
  \captionsetup{justification=centering}
  \caption{Backward \& In the Middle}
  \label{fig_backward_in_the_middle}
\end{subfigure}
\caption{Text Generation}
\label{fig:text_generation_2}
\end{figure}

In Fig.~\ref{fig_forward_in_the_middle}, row (1) shows the same tokens of the input text. Row (2) indicates that the calculation starts from \emph{$t_n$} in the middle (\emph{n} can be any number between 0 and m) and is about to move forward. Row (3) indicates that the calculation moves forward from \emph{$t_n$}, reaches \emph{$t_m$} (the end), and is about to move backward. For moving backward, row (3) is reversed as row (4). The calculation now moves forward starting from \emph{$t_n$} again. Given \emph{$t_m$}, \emph{$t_{m-1}$}, ..., \emph{$t_{n+1}$}, \emph{$t_n$} as the context, row (5) indicates that the calculation moves forward and reaches \emph{$t_0$} (the beginning). Row (6) shows the reversed sequence of row (5) after moving forward and then backward. 

In Fig.~\ref{fig_backward_in_the_middle}, row (1) shows the same tokens of the input text. Row (2) indicates that the calculation starts from \emph{$t_n$} in the middle (\emph{n} can be any number between 0 and m) and is about to move backward. Row (3) is the reversed sequence of row (2) so that the calculation moves backward in effect. Row (4) indicates that the calculation moves forward from \emph{$t_n$} and reaches \emph{$t_0$}.
For moving forward with respect to the original input, row (4) is reversed as row (5). The calculation continues moving forward and starts from \emph{$t_n$} again. Given \emph{$t_0$}, \emph{$t_1$}, ..., \emph{$t_{n-1}$}, \emph{$t_n$} as the context, row (6) indicates that the calculation moves forward and reaches \emph{$t_m$} as the end of the input text.

\subsection{Further Fine-tunig}
\label{subsection:fine_tuning}

Another objective of this manuscript is to see how the AE ratio may improve if the model is fine-tuned with a specific category of patents. The specific category in this section is the CPC Subclass G06N. Other categories could be experimented with in the future. According to the CPC definition, the subclass G06N covers computing systems where the computation is not based on a traditional mathematical model of computer. For example, G06N covers machine learning, neural network models, knowledge-based models, artificial intelligence, etc. G06N is relevant to the technical field in this manuscript. If there is any patentable idea in this manuscript, the idea should fall within the same subclass. 




\subsection{Data}
\label{subsection:data}

The dataset in this manuscript is the USPTO TACD dataset in~\cite{jiehsheng_n01} which includes titles, abstracts, claims, and descriptions in granted patents. The coverage of the dataset ranges from 1976 to 2021 (1976$\sim$2020 for training and 2021 for validation). The dataset covers both independent claims and dependent claims. Dependent claims are expanded with their independent claims. For generating patent text forward, the dataset contains the tokens of patent text after tokenization. Most generative language models generate text forward. For generating patent text backward, a copy of reversed tokens of all patent text is pre-processed and added to the dataset. After expanding the dependent claims and adding a copy of the reversed tokens, the total size of the USPTO TACD dataset in plain text is about 713 GiB. Following the same practices in~\cite{jiehsheng_n01}, pre-training a PatentGPT-J model in this manuscript consumes about 7.48\% of the USPTO TACD dataset (11B tokens out of 147B) after training 350,000 steps. For further details such as tokenizer, additional items in the vocabulary file, and how patent claim expansion works, please refer to~\cite{jiehsheng_n01}.



\subsection{Model sizes \& Training Losses}
\label{subsection:model_size}

\begin{figure}[h]
  {\includegraphics[width=0.8\textwidth, keepaspectratio]{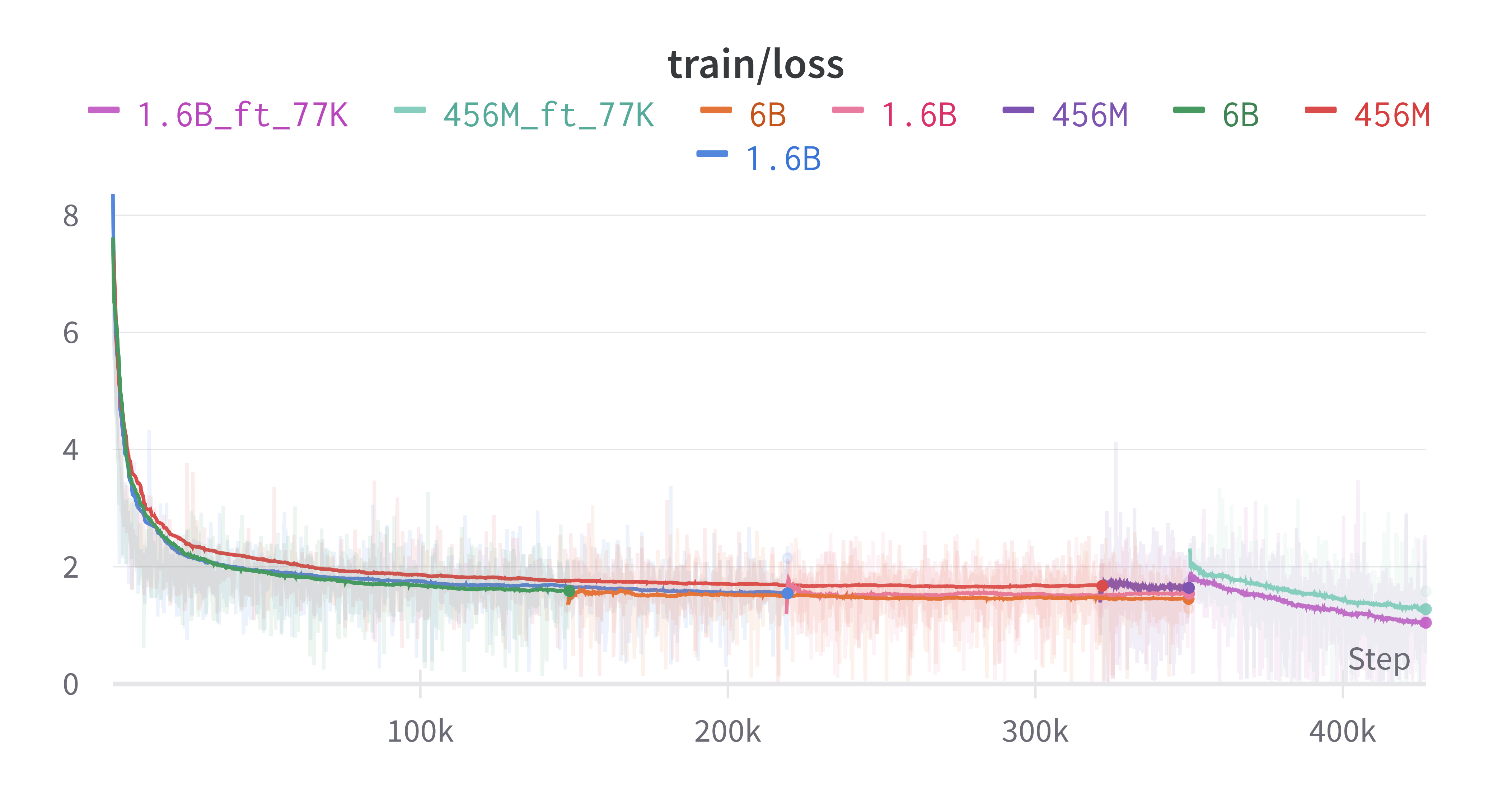}}
  \caption{Training loss curve}
  \label{training_loss}
\end{figure}

The model sizes experimented in this manuscript are 6B, 1.6B, and 456M. These three model sizes outperform others (279M, 191M, 128M, and 115M) in~\cite{jiehsheng_n01}. The training losses of the three models in this manuscript are shown in Fig.~\ref{training_loss}. The curves ``1.6B\_ft\_77K'' and ``456M\_ft\_77K'' represent the models fine-tuned with 77K steps. The fine-tuning stage is conducted after building the pre-trained models. It is noted that the loss curve at the fine-tuning stage dives deeper than the curve at the pre-training stage. However, the AE ratio based on the fine-tuned model is not significantly better than that of the pre-trained model. Details will be discussed in~\ref{subsection:experiment_4}. 
In~\cite{jiehsheng_n01}, the models are released at the repository~\cite{jiehsheng_PatentGPT-J_github}. After publication, the pre-trained and fine-tuned models in this manuscript will be released. The model configuration in this manuscript was reused from~\cite{jiehsheng_n01}. In terms of training from scratch, this manuscript uses the original GPT-J-6B model and its source code available at~\cite{gpt-j-github}. 

\section{Experiments} 
\label{section:experiments}

\subsection{Experiment 1}
\label{subsection:experiment_1}

This experiment aims to compare the effectiveness of the Algorithm~\ref{alg:counting_keystrokes_v2} in this manuscript with the previous Algorithm in~\cite{jiehsheng_n01}. The increase in effectiveness in this manuscript is more than 10\%, as shown in the fifth column of Table~\ref{table:evaluation1}. The first column shows the three models in this experiment: 456M, 1.6B, and 6B. The second column shows the direction for calculating the AE ratio: forward and backward. The third column shows the AE ratios of using the previous Algorithm in~\cite{jiehsheng_n01}. The fourth column shows the new AE ratios based on Algorithm~\ref{alg:counting_keystrokes_v2} in this manuscript. The ratio of more than 62\% in the fourth column means that more than 62\% of keystrokes can be saved by autocomplete in this experiment. The effectiveness of the simpler autocomplete function using keys 0$\sim$9 is validated. The previous combination of the ``downarrow'' ($\downarrow$) key and the ``tab'' key in~\cite{jiehsheng_n01} is eliminated. In Table~\ref{table:evaluation1}, the ``Test Data A'' are the patent data used in the first experiment of~\cite{jiehsheng_n01} and reused here. The patent data cover 500 independent claims randomly selected from patents in 2022.

\begin{table}[h]
  \begin{tabular}{c c c c c}
    Model & Direction & Previous & New & Increase ($\uparrow$) \\ 
     & & Ratio ($\uparrow$) & Ratio ($\uparrow$) & \\ \hline\hline
    456M & forward & 56.5\% & 62.7\% & 10.9\% \\ \hline
    456M & backward & 55.7\% & 62.1\% & 11.4\% \\ \hline
    1.6B & forward & 57.0\% & 63.1\% & 10.7\% \\ \hline
    1.6B & backward & 56.2\% & 62.5\% & 11.2\% \\ \hline
    6B & forward & 57.0\% & 63.1\% & 10.7\% \\ \hline
    6B & backward & 56.5\% & 62.7\% & 10.9\% \\ \hline \\
  \end{tabular} 
\caption{The improvement of the AE ratio. Target: Test Data A.}
\label{table:evaluation1}
\end{table}



\subsection{Experiment 2}
\label{subsection:experiment_2}

Table~\ref{table:evaluation2} shows another comparison between the Algorithm~\ref{alg:counting_keystrokes_v2} in this manuscript and the previous Algorithm in~\cite{jiehsheng_n01}. Compared to Table~\ref{table:evaluation1}, the only difference in this experiment is the patent data used to calculate the AE ratios. In Table~\ref{table:evaluation2}, the ``Test Data B'' are the patent data used in the second experiment of~\cite{jiehsheng_n01} and reused here. The patent data cover another 500 patent claims randomly selected from 2022 and include independent and dependent claims. The description of the table structure is the same and is omitted here for brevity. It is noted that the ratios in the fourth column are slightly lower than the ratios in Table~\ref{table:evaluation1}. However, the new AE ratios are still around 60\% in this experiment. Such results implies the stability of model performance for the autocomplete function. 

\begin{table}[h]
  \begin{tabular}{c c c c c}
    Model & Direction & Previous & New & Increase ($\uparrow$) \\ 
     & & Ratio ($\uparrow$) & Ratio ($\uparrow$) & \\ \hline\hline   
    456M & forward & 54.2\% & 60.0\% & 10.7\% \\ \hline
    456M & backward & 54.3\% & 58.9\% & 8.4\% \\ \hline
    1.6B & forward & 54.7\% & 60.1\% & 9.8\% \\ \hline
    1.6B & backward & 54.0\% & 59.5\% & 10.1\% \\ \hline
    6B & forward & 55.0\% & 60.6\% & 10.1\% \\ \hline
    6B & backward & 54.3\% & 59.6\% & 9.7\% \\ \hline \\
  \end{tabular} 
\caption{The improvement of the AE ratio. Target: Test Data B.}  
\label{table:evaluation2}
\end{table}


\subsection{Experiment 3}
\label{subsection:experiment_3}

This experiment is similar to the previous two. The only difference is in using another set of patent claims for AE calculation. The ``Test Data C'' in this experiment contains 1,000 patent claims of CPC Subclass G06N. These patents have not been used previously in\cite{jiehsheng_n01}. Table~\ref{table:evaluation3} shows the experiment results. Overall the new AE ratios are more than 60\%. The effectiveness of Algorithm~\ref{alg:counting_keystrokes_v2} increases over 10\%, compared to its previous version in~\cite{jiehsheng_n01}, as shown in the fifth column. Besides calculating the AE ratios based on the pre-trained models, the other purpose of using these G06N patents is to compare the AE ratios based on fine-tuned models with G06N patents later. In this experiment, the models are pre-trained models. The models in the next experiment are fine-tuned.
The description of the table structure is the same and omitted. 

\begin{table}[h]
  \begin{tabular}{c c c c c}
    Model & Direction & Previous & New & Increase ($\uparrow$) \\ 
     & & Ratio ($\uparrow$) & Ratio ($\uparrow$) & \\ \hline\hline   
    456M & forward & 55.1\% & 60.9\% & 10.5\% \\ \hline
    456M & backward & 54.0\% & 60.0\% & 11.1\% \\ \hline
    1.6B & forward & 55.5\% & 61.3\% & 10.4\% \\ \hline
    1.6B & backward & 54.5\% & 60.5\% & 11.0\% \\ \hline
    6B & forward & 55.7\% & 61.3\% & 10.0\% \\ \hline
    6B & backward & 54.6\% & 60.6\% & 10.9\% \\ \hline \\
  \end{tabular}
\caption{The improvement of the AE ratio. Target: Test Data C.}
\label{table:evaluation3}
\end{table}



\subsection{Experiment 4}
\label{subsection:experiment_4}

This experiment aims to compare the AE ratios of the pre-trained models with the fine-tuned models. The target data to calculate the AE ratios is the ``Test Data C'' described in~\ref{subsection:experiment_3}. Table~\ref{table:evaluation4} shows the experiment results. Before the experiment, it is expected that a fine-tuned model will perform much better than its pre-trained counterpart. The result in Table~\ref{table:evaluation4} shows otherwise. Regarding the 456M model, the improvement from 60.9\% to 62.9\% for the forward direction is not significant, as shown in the first row and the second row in Table~\ref{table:evaluation4}. The improvement of the 1.6B model is even smaller (61.3\% to 62.0\%, third row \& fourth row). The 6B model is omitted since the improvement is unlikely to be significant. How to fine-tune a model more effectively and reach a higher AE ratio is a research topic in the future. 

\begin{table}[h]
  \begin{tabular}{c c c}
    Model & Forward & Backward\\ \hline\hline    
    456M (w/o fine-tuning) & 60.9\% & 60.0\% \\ \hline
    456M (w/ fine-tuning) & 62.9\% & 61.4\% \\ \hline
    1.6B (w/o fine-tuning) & 61.3\% & 60.5\% \\ \hline
    1.6B (w/ fine-tuning) & 62.0\% & 60.4\% \\ \hline \\
  \end{tabular}
\caption{Comparison of AE ratios after fine-tuning. Target: Test Data C.}  
\label{table:evaluation4}
\end{table}



\subsection{Experiment 5}
\label{subsection:experiment_5}

This experiment implements the mechanism of moving back and forth described in section~\ref{subsection:back_and_forth}. Table~\ref{table:evaluation5} shows the experiment results. In the table, the third column ``Q1'' means the position of the 25\%-th token of the input text. The fourth column ``Q2'' means the position of the 50\%-th token of the input text. The fifth column ``Q3'' means the position of the 75\%-th token. The second column defines the direction for calculating the AE ratios. For example, if the input text is tokenized and has 100 tokens, the position ``Q1'' means that the calculation starts from the 25th token. The ``forward'' direction in the second column means that the calculation of the AE ratio moves forward to the 26th, 27th, ..., 100th tokens. The 100th token is the end of the input text. After reaching the end, the calculation starts from the 25th token again. The calculation then moves in effect backward to the 24th, 23th, ..., 1st tokens. The forth-and-back calculation in this example runs over all 100 tokens. If the direction is ``backward,'' the calculation will move back first and forth later to run over all 100 tokens. According to Table~\ref{table:evaluation5}, the AE ratios are similar to one another no matter where the starting position is and no matter which direction to go first. Such a finding indicates that, no matter where a user starts to write, the autocomplete function based on PatentGPT-J models can assist the user to a similar degree and save a similar number of keystrokes. This manuscript hypothesizes that such a property is not specific to the patent domain and may apply to other generative language models with different training data. This hypothesis will be validated in the future.

\begin{table}[h]
  \begin{tabular}{c c c c c}
    Model & Direction & Q1 & Q2 & Q3 \\ \hline\hline    
    456M (fine-tuned) & forward & 62.2\% & 61.7\% & 61.3\% \\ \hline
    456M (fine-tuned) & backward & 62.4\% & 62.1\% & 61.7\% \\ \hline
    1.6B (fine-tuned) & forward & 61.8\% & 60.9\% & 60.1\% \\ \hline
    1.6B (fine-tuned) & backward & 61.6\% & 61.4\% & 60.9\% \\ \hline \\
  \end{tabular}
\caption{The AE ratios from different starting positions. Target: Test Data C.}  
\label{table:evaluation5}  
\end{table}

\section{Conclusion}
\label{section:conclusion}

Generative language models have great potential to assist humans in writing patent text more effectively. In this manuscript, the way to measure effectiveness is to calculate the ratio of keystrokes saved by the autocomplete function based on model predictions. A higher ratio means more saved keystrokes and fewer manual typing. The ratio was proposed in previous work as the AE ratio. After using a simplified design in this manuscript, the AE ratio increases by more than 10\% and reached more than 60\%. The new AE ratio of more than 60\% means that more than 60\% of keystrokes can be saved by using the autocomplete function.
Furthermore, the generative models can go forward and backward, after pre-training with training data in both directions. The models are bidirectional. Such bidirectional models make it possible to calculate the AE ratio in both directions. The calculation can start anywhere in the text. A key finding is that the AE ratio for the same text remains similar regardless of where the calculation starts. This finding indicates that such bidirectional models can assist a user at a similar level, no matter where the user starts to write. In addition to the patent domain, the research in this manuscript can be applied to other legal domains in the future because the Transformer architecture in generative models is domain-agnostic. 

\section{Acknowledgments}
The research reported in this manuscript has been funded by the
Ministry of Science and Technology (MOST) in Taiwan (Project ID:
111-2222-E-A49-005). In addition, the author would like to thank TensorFlow Research Cloud (TRC) greatly for providing significant computational resources. Without the resources, it would not be easy to build several language models efficiently to cover a range of implementations (pre-training, fine-tuning, inferencing, calculating the Autocomplete Effectiveness ratio, etc.).

\bibliographystyle{vancouver.bst}
\bibliography{citation}

\end{document}